\documentclass[11pt,a4paper]{article}
\usepackage[hyperref]{emnlp2020}
\usepackage{times}
\usepackage{latexsym}
\renewcommand{\UrlFont}{\ttfamily\small}

\usepackage{microtype}
\usepackage{multirow}
\usepackage{graphicx}
\aclfinalcopy % Uncomment this line for the final submission
\title{\texttt{doc2dial}: A Goal-Oriented \textbf{Doc}ument-Grounded \textbf{Dial}ogue Dataset }

\author{
Song Feng \quad Hui Wan  \quad Chulaka Gunasekara \quad Siva Sankalp Patel \\ 
{\bf Sachindra Joshi \quad  Luis A. Lastras}  \\
IBM Research AI \\
\texttt{\{sfeng@us, hwan@us, chulaka.gunasekara@\}.ibm.com} \\
\texttt{\{siva.sankalp.patel@, jsachind@in, lastrasl@us\}.ibm.com}
}

\date{}

\begin{document}
\maketitle
\begin{abstract}
We introduce \textbf{\texttt{doc2dial}}, a new dataset of goal-oriented dialogues that are grounded in the associated documents. Inspired by how the authors compose documents for guiding end users, we first construct dialogue flows based on the content elements that corresponds to higher-level relations across text sections as well as lower-level relations between discourse units within a section. Then we present these dialogue flows to crowd contributors to create conversational utterances. The dataset includes about 4800 annotated conversations with an average of 14 turns that are grounded in over 480 documents from four domains. Compared to the prior document-grounded dialogue datasets, this dataset covers a variety of dialogue scenes in information-seeking conversations. For evaluating the versatility of the dataset, we introduce multiple dialogue modeling tasks and present baseline approaches. 
\end{abstract}

\section{Introduction}

The task of reading documents and responding to queries has been the trigger of many recent research advances. On top of the development of contextual question answering QuAC~\citep{choi2018quac} and CoQA~\citep{reddy2019coqa}, more recent work MANtIS~\citep{penha2019mantis} and DoQA~\citep{campos-etal-2020-doqa} included more kinds of user intents for querying over documents; while ShARC~\citep{saeidi2018interpretation} added follow-up questions from agents and binary answers from users for the inference over documents. These exciting works confirm the importance of modeling document-grounded dialogue. Yet, it involves more complex scenes in practice, which requires better understanding of the inter-relations between conversations and documents. Thus, we aim to investigate how to create the training instances to further approach real-world applications of document-grounded dialogue for information seeking tasks.

\begin{figure*}[!t]
\centering
\includegraphics[width=0.9\textwidth]{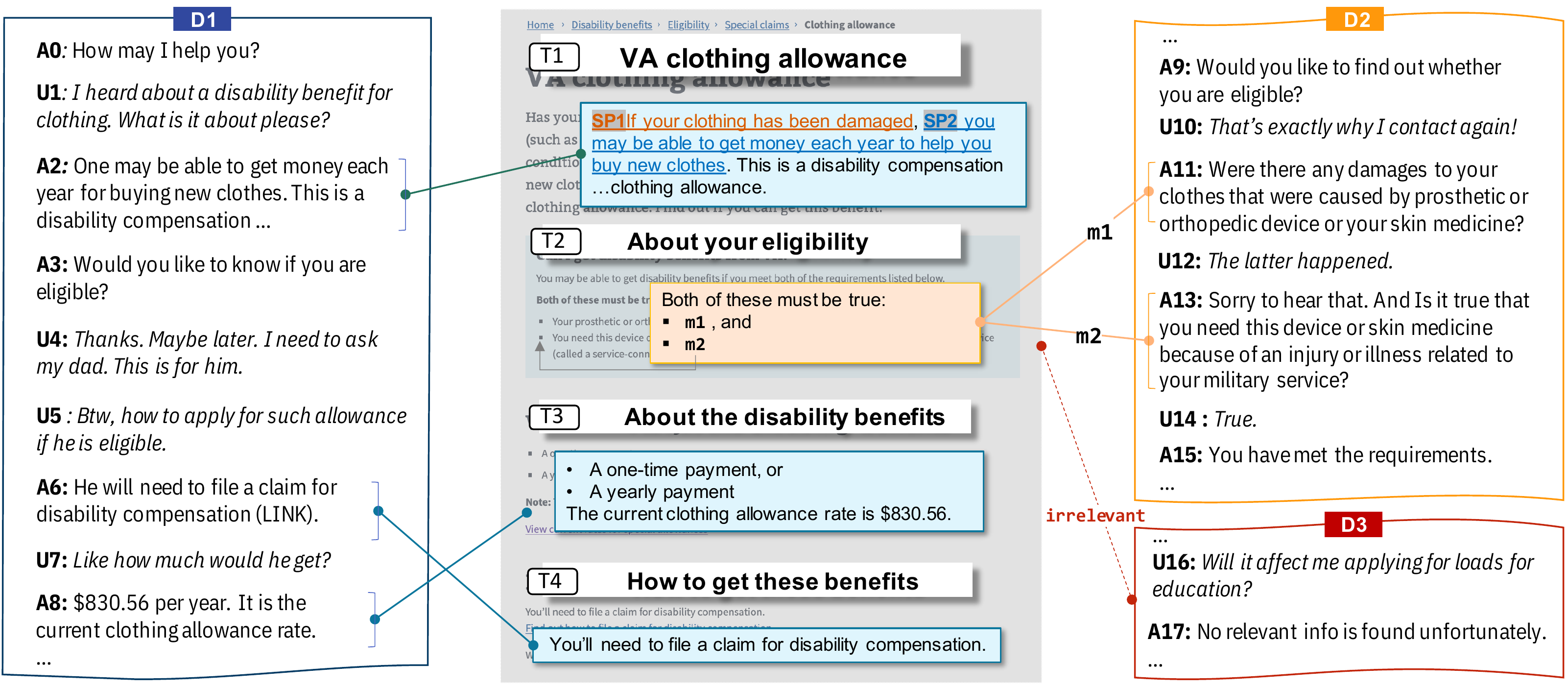}
\caption{\label{fig:intro} Sample segments of conversations (\texttt{D1}, \texttt{D2} and \texttt{D3}) with various dialogue scenes that are grounded in a webpage (middle) from {\UrlFont{va.gov}}. 
The relevant content elements, such as hierarchical headers, list-items and spans, are highlighted. 
\texttt{A} / \texttt{U} indicates \underline{A}gent / \underline{U}ser role. }
\end{figure*}

In this work, we propose a new dataset of goal-oriented document-grounded dialogue. Figure \ref{fig:intro} shows sample utterances from dialogues \texttt{D1}, \texttt{D2} and \texttt{D3} between an assisting agent and a user, and an example document in the middle. \texttt{D1} and \texttt{D2} are grounded in the given document, while \texttt{D3} is irrelevant to the document. It illustrates two different types of contexts that we aim to capture: (1) \emph{dialogue}-based context, where a query could be formed by a single or multiple turns, and (2) \emph{document}-based context, which corresponds to varied forms of knowledge represented in the document. More specifically, \emph{dialogue}-based context of a query could be initiated by a user (e.g., {U1} in \texttt{D1}) or an agent (e.g., {A3} in \texttt{D1}), and carried out through multiple turns by both roles (e.g., all turns in \texttt{D2}). \emph{Document}-based context could involve structural elements in documents, such as the headers \texttt{T1} and \texttt{T2} or list items of \texttt{m1} and \texttt{m2}, as well as textual discourse units, such as clauses (e.g, ``If your clothing has been damaged'').

For creating such dataset, we consider the document contents for social welfare websites,such as {\UrlFont ssa.gov} and {\UrlFont va.gov}, which guide users to access various forms of information. We develop a pipeline approach for dialogue data construction. Inspired by how human authors compose user-facing web content, we utilize both the high-level hierarchical relations between document components, as well as the low-level semantic relations between discourse units \citep{stede2019connective} to dynamically create outlines of dialogues, or we call dialogue flows. A \emph{\textbf{dialogue flow}} is a sequence of interactions between an assisting agent and a user. Each turn contains a \emph{\textbf{dialogue scene}} that is defined by a dialogue act, a role (user or agent) and a piece of grounding content from a document. Then we present these dialogue flows to crowd contributors to create conversational utterances. Such approach helps to avoid additional noise from the post-hoc human annotations of dialogue data, which is a known challenge \citep{geertzen2009measuring}.

The dataset contains about 4800 annotated conversations with an average of 14 turns per dialogue. The utterances are grounded in over 480 documents from four domains. Unlike the previous work on document-grounded question answering or dialogues \citep{choi2018quac,reddy2019coqa,saeidi2018interpretation} that are based on a short text snippet, our dialogues are grounded in a much wider span of context in the associated documents. 

For evaluation, we propose three tasks that are related to identifying and generating responses with grounding content in documents: (1) user utterance understanding; (2) agent response generation; and (3) relevant document identification. For each task, we present baseline approaches and evaluation results. Our goal is to elicit further research efforts on building document-grounded dialogue models that can incorporate deeper contexts for tackling goal-oriented information-seeking tasks. 
We summarize our main contributions as follows:
\begin{itemize}
    \item We introduce a novel dataset for modeling dialogues that are grounded in documents from multiple domains. The dataset is available at \url{http://doc2dial.github.io/}.
    \item We develop a pipeline approach for dialogue data collection, which has been adapted and evaluated for varied domains.
    \item We propose multiple dialogue modeling tasks that are supported by our dataset, and present the baseline approaches.
\end{itemize}

\section{Doc2Dial}
We introduce \textbf{\texttt{doc2dial}}, a new dataset that includes (1) a set of documents; and (2) conversations between an assisting agent and an end user, which are grounded in the associated documents. Figure \ref{fig:intro} presents sample utterances from different dialogues along with a sample document from {\UrlFont va.gov} in the middle. It illustrates some prominent features in our dataset, such as the cases where a conversation involves multiple interconnected sub-tasks under a general inquiry (e.g., \texttt{D1}); or the cases where a conversation involves multiple interactions to verify the conditional contexts for one query (e.g., \texttt{D2}).

Recent work, such as \citet{saeidi2018interpretation}, has started to address the challenge of modeling complex contexts by allowing follow-up questions from agents based on natural language inference rules extracted from the relevant documents. However, it also simplified the task by using only restricted forms of questions and binary answers. In our work, we not only encourage free-form utterances, but also aim to include various dialogue scenes that provoke inquires with different \emph{document}-based and \emph{dialog}-based contexts. A user query can be formed in single-turn or multiple-turn manners: (1) the user explicitly states a context that is associated with a text-span that contains a solution to the query, e.g., U5 on T4; (2) the user describes an implicitly stated context associated with a solution, e.g., U7; (3) the user accepts or rejects a piece of agent-stated context that is associated with a solution, e.g., U4 (rejection), and U12 \& U14 (acceptance). 
An agent response, on the other hand, either provides a solution or poses a query depending on the context of a given user query: (1) whether the query is irrelevant to the grounding document, e.g., A17; (2) whether the query is under-specified, if so, the agent will suggest associated context, e.g., A11 and A13; (3) whether a relevant answer is identified in the grounding document, e.g, A6, A8 and A15. 

\begin{figure}
\centering
\includegraphics[width=0.5\textwidth]{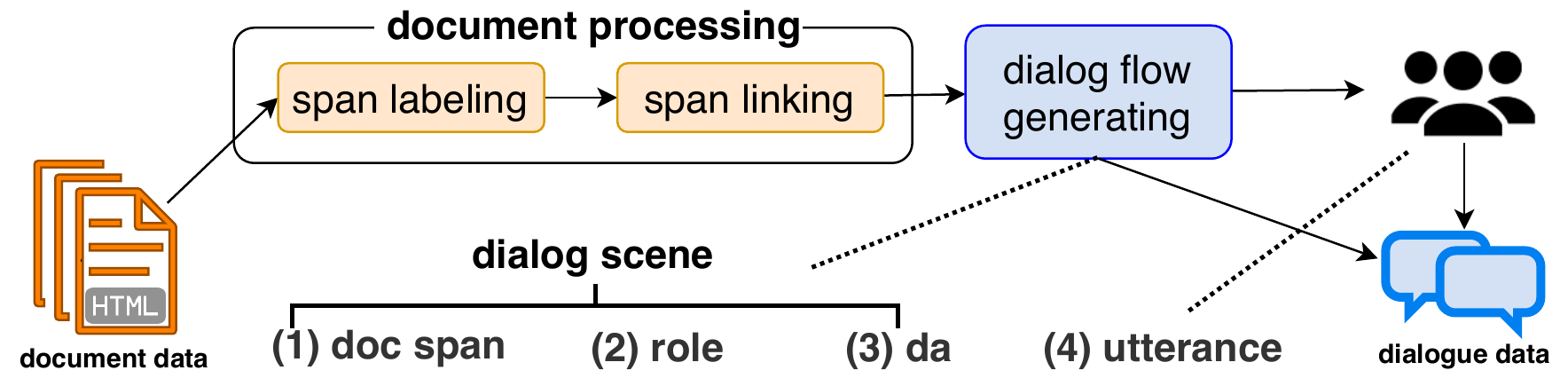}
\caption{\label{fig:sys} The overview of the process for constructing and annotating \textbf{\texttt{doc2dial}} dataset.}
\end{figure} 

\subsection{Data Collection}
For collecting document-grounded dialogue data, we propose a pipeline approach derived from the framework proposed by \citet{feng2019doc2dial}. As shown in Figure \ref{fig:sys}, it includes the components for: (1) processing the document contents; (2) generating dynamic dialogue flows; (3) crowdsourcing the dialogue utterances. 

\subsubsection{Data Construction Approach}

\paragraph{Processing document contents} We first select documents that contain the context-indicative elements, such as hierarchical headers and explicit discourse relations \citep{prasad-etal-2008-penn,prasad2019penn}, since those document contents could provoke more diversified dialogue flows. Then we extract text-spans to create a graph with the spans as nodes and semantic relations as edges. Some spans in the graph correspond to a piece of information for solving user problems, while some correspond to the conditional context of those solutions, such as \texttt{SP2} and \texttt{SP1} in Figure \ref{fig:intro} respectively. The semantic relations are largely determined by the heuristics derived from the document structures \citep{mukherjee2003automatic} and semantic connectives \citep{das2018constructing} between discourse units or clauses. Both spans and semantic relations are labeled automatically via our tool. The labels can be reviewed and annotated via crowdsourcing platforms, which is also supported by our tool.
% Some illustrative examples are included in Appendix A.

\paragraph{Generating dynamic dialogue flows} Each flow consists of a sequence of dialogue scenes. A \emph{\textbf{dialogue scene}} is described with (1) role, either a user or an agent; (2) a selected span as the grounding content from the given document; (3) a dialogue act that determines how to describe the selected span in the given role. Thus, each turn is inherently annotated with the dialogue act and a reference to the document contents. The dynamics of the dialogue flows are introduced by varying the three factors that are constrained by the relations from the semantic graph and dialogue history. In principle, we randomly select content from a candidate pool of spans of conditional contexts and solutions. The pool is updated after every turn is generated based on the status of the previously selected span. The general rule for updating the candidate pool is to avoid re-selecting any spans with an established status. In addition, the dialogue flow is principally aligned with common practice of dialogue management, for instance, after an agent asks a user a question, we expect the next turn would be the user answering the question.
% The details of creating dialogue flows from the span graph are included in Appendix A.

\paragraph{Collecting human utterances} Finally, we present the sequences of dialogue scenes to crowdsourced contributors to convert them into conversational utterances.
% we generate the crowdsourcing tasks for collecting the utterances in English based on the given dialogue scene and dialogue history. We also collect additional feedback if a writer finds the dialogue scene not feasible for creating the next coherent conversation turn. 

\subsubsection{Crowdsourcing Setup} 
Our data collection task asks the crowd contributors to focus on one turn at a time so that they can carefully review the given dialogue scene and the dialogue history. Since the crowd generally prefers to work on tasks in batches, we try different settings to combine the tasks: (1) each writer plays the same role but for different dialogues per batch; or (2) each writer plays both agent and user role and completes entire dialogue in order, as inspired by \citet{byrne2019taskmaster}. We also find that the conversations by the second setting tend to be more coherent and less time consuming. Many writers would make efforts to differentiate their writing styles for different roles. Therefore, our tasks were completed based on the second setting by about 70 qualified contributors from {\UrlFont{appen.com}}. We pay \$1.5-\$2 per conversation. 

\begin{table}[!t]
    \centering
    \small
    \begin{tabular}{l r | r |rrrr } \\ \hline
     \multirow{2}{*}{\textbf{Domain}} & \multirow{2}{*}{\textbf{\#Dials}} & \multirow{2}{*}{\textbf{\#Docs}} & \multicolumn{4}{c}{\textbf{\# per doc}} \\  
     &  &  &  \textbf{tk} & \textbf{sp} & \textbf{p} & \textbf{sec}  \\ \hline
    % \texttt{ssa.gov}  & 860     & 86     & 758  & 66 & 16 & 5     \\
    % \texttt{va.gov}   & 1340    & 138    & 823  & 70 & 20 & 9   \\
    % \texttt{dmv.gov}  & 1420    & 149    & 955  & 77 & 18 & 10 \\
    % \texttt{cdc.gov}  & 850     & 85     & 1251 & 94 & 16 & 9    \\ \hline
    % \texttt{all}      & 4470    & 458    & 947  & 77 & 18 & 8 \\ \hline
    % new numbers
ssa & 1192 & 109 & 795 & 70 & 17 & 5 \\
va & 1330 & 138 & 818 & 70 & 20 & 9 \\
dmv & 1305 & 149 & 944 & 77 & 18 & 10 \\
studentaid & 966 & 91 & 1007 & 75 & 20 & 9 \\ \hline
all & 4793 & 487 & 888 & 73 & 18 & 8 \\ \hline
\end{tabular}
\caption{The breakdown count of the dialogues, documents and average number of content elements per document by domain.}
\label{tb:datastats}
\end{table}

\subsection{Document Data} 
For document contents, we consider the public government service websites that are designated to provide information to a vast group of users. We collect web contents from four domains and select over 480 documents for creating dialogue flows as shown in Table \ref{tb:datastats}. Our dataset provides document contents in plain text and HTML, along with the meta information of titles and URLs. Each document is also represented as a sequence of spans, for which we provide indexes to the plain text and the HTML respectively. 
% In particular, we exploit certain shared writing style across these websites, which includes how the human authors employ the higher-level relations between document components such as hierarchical headers and semantically diversified list-items via HTML markups; as well as the lower-level relations, such as ``is reason of'' or ``exclusive or'', between clauses or discourse units. 
% , such as the examples below where the italic text indicates the conditional statements implicitly (a) and explicitly (b).
% for the purpose of explaining the contexts and corresponding solutions for end users. 

% \begin{quote}
% \small{
%     (a): Please follow LINK \textit{to file a claim}. \\
%     (b): \textit{If you need to file a claim}, go to LINK.
%     }
% \end{quote}

\paragraph{Content elements} To characterize the document contents, we examine the HTML source to extract the content elements with different scopes such as, tokens (\textbf{tk}), spans (\textbf{sp}), paragraphs (\textbf{p}) and titled sections (\textbf{sec}). Some of the spans within one sentence, such as \texttt{SP1} and \texttt{SP2} in Figure \ref{fig:intro}, are extracted via constituency parsers \citep{joshi2018extending}. The paragraphs and sections are determined using HTML markups. The average counts of these elements per document in Table \ref{tb:datastats} show the rich structures that are employed across domains. While this work starts to explore the simpler semi-structured information such as D2 in Figure \ref{fig:intro}; we are yet to explore various semantics from complex list structures, tables and other multi-modal contents in the webpages for future work.

\begin{table}[!t]
    \centering
    \small
    \begin{tabular}{l l r r} \\ \hline
    \textbf{Role} & \textbf{DA} & \textbf{\#Turns} & \textbf{\#Tokens/Turn} \\  \hline
    user & request/query & 22220 & 12 \\
    user & respond/yesOrNo & 8413 & 6 \\
    agent & request/query & 7927 & 12 \\ \hline
    agent & respond/reply & 23482 & 21 \\
    total &  all    & 62042 & 14 \\ \hline
    \end{tabular}
    \caption{The total \# of turns and the average \# of tokens per turn, aggregated on dialogue act category.} 
    % hierarchical  
    \label{tab:dialstats}
\end{table}

\subsection{Dialogue Data} 
Given a grounding document, we create about multiple unique dialogue flows with an average of 14 turns for this dataset. All dialogues are created based on a unique dialogue flow. In total, there are close to 4800 conversations with about 62,000 turns from over 480 document in four domains as shown in Table \ref{tb:datastats}. Each dialogue utterance is annotated with a dialogue scene, i.e., role, dialogue act and the grounding span. As it is a known challenge to annotate conversation turns for the dialogue scenes \citep{geertzen2009measuring}, our pipeline approach for data collection helps avoid the cost and the noise from the additional human annotations. Next we further describe it from different perspectives regarding the dialogue scene.

\paragraph{Dialogue acts}  We adopt the hierarchical dialogue act scheme by \citet{pareti2018dialog} with a focus on the ones most essential to the information-seeking tasks. We describe those dialogue acts to the crowdsourced contributors pertaining to the selected grounding content and the assigned role (detailed descriptions in Appendix A). For future work, we plan to extend current dialogue scenes with other actions such as elucidations \citep{azzopardi2018conceptualizing} and social acts \citep{kluwer2011like}. To examine the dialogue distributions, we aggregate the hierarchical dialogue acts and list the total of turns, and the average length per turn under each category in Table \ref{tab:dialstats}. For example, ``agent | request/query'' corresponds to the queries based on document-guided dialogue management turns via an agent role; ``user | respond/yesOrNo'' corresponds to the scene where a user responds to an agent's query. Since we encourage the crowd to express ``yes'' or ``no'' in natural and creative writings, such as U10 in D2 in Figure \ref{fig:intro}, the average length of ``respond/yesOrNo'' is 7 tokens.
%In this datasets, most of the user turn corresponding to information-seeking, such as specifying the context of the query, the query that can be answered with document contents or cannot be answered with the grounding document. Currently, for the agent prompts, we focus on the dialogue acts on information verification determined by the document-based content. For instance, as illustrated in Figure \ref{fig:intro}, agent turn \texttt{A} and \texttt{A} are the two-step verification for the end users' eligibility. User turn \texttt{U} asks a contextual question and agent turn \texttt{A} provides the precise answer when it applies. 

\begin{figure}[!t]
\centering
\includegraphics[width=0.50\textwidth]{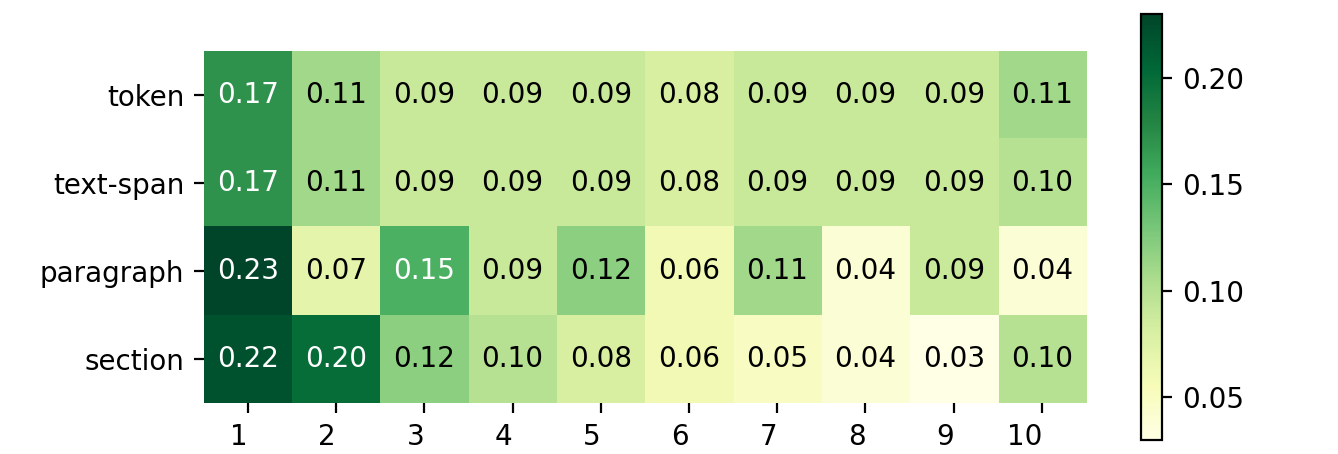}
\caption{\label{fig:index} An illustration of the indexes of the relevant grounding contents in the documents.}
\end{figure}

\paragraph{Grounding content} We aim to include the contents that are associated with varied conditional contexts based on the aforementioned span graph without introducing strong bias on certain index position in the document as discussed in \citet{geva-berant-2018-learning}. Therefore, we examine the coverage of the document contents from the generated dialogue flows. As illustrated in Figure \ref{fig:index}, we create index of all the selected grounding contents to different document segments such as tokens, spans, paragraphs and titled sections (y-axis). The x-axis (numbered 1-10) indicates the position where 1 is closest to the beginning and 10 is closest to the end of a document. The numbers in the cells indicate the percentage distribution among all the grounding contents. The heatmap shows some degree of coverage on all parts of the documents, with a higher density at the beginning as we do include the scenarios of under-specified queries that typically correspond to the intro of a document.

\begin{table}
    \centering
    \small
    \begin{tabular}{p{0.36\textwidth}p{0.05\textwidth}} 
    % \begin{tabular}{ll}
    \hline
    \textbf{feedback on rejected dialogue scene} & \textbf{\%} \\ \hline
    The selected-text is not a contextual condition. & 74.3 \\
    The selected-text is not a solution to the query. & 10.5 \\
    Cannot write a turn to be coherent with the chat history. & 10.1 \\
    There is not enough information in the selected (or adjacent) text. & 2.4 \\
    The selected-text is not Comprehensible. & 1.8 \\
    Other. & 0.9 \\ \hline
\end{tabular}
\caption{Feedback on the reasons for rejecting a dialogue scene by crowdsourced annotators.}\label{tb:feedback}
\end{table}

\paragraph{Dialogue flows} For assessing the quality of the dialogue flows, we also ask the contributors to reject a dialogue turn when it is considered as infeasible to write a coherent utterance. We also solicit feedback via multiple choices on the reason as shown in Table \ref{tb:feedback}. Out of 700 sampled dialogue flows, annotators reject about 4\% of the turns. Among the rejected turns, 70\% is due to not being able to interpret the selected span as applicable conditional context for user requests. In this dataset, we exclude the (sub)dialogues with rejected turns accordingly. However, we also observe certain ``false positive'' cases, where the crowd would rather try to adjust their writing for a less desirable dialogue scene rather than rejecting the turn, for which they get paid the same either way.

\subsection{Data Recomposition} 
\label{ss:recomp}
One benefit of constructing the dialogue data via our pipeline approach is that it provides a convenient and cost-effective way to reshape the existing dialogue data based on their dialogue flows.
% for quality improvement or model training purpose. 
For instance, to ensure the quality, we can recollect or remove certain turns from the dialogues if they are rejected by the crowd contributors or affected by the changes in the grounding documents. In addition, for obtaining the training instances to identify the irrelevant queries, we modify an existing dialogue by inserting sub-dialogues created for another document or domain, for instance, adding \texttt{D3} to \texttt{D1} as \textit{irrelevant} for {\UrlFont{va.gov}} in Figure \ref{fig:intro}. Similarly, for creating dialogues that are grounded in multiple documents, we select sub-dialogues based on different documents and combine them into one.

\section{Tasks and Baselines}
For evaluation, we propose three tasks related to identifying the grounding content for a given dialogue: (1) user utterance understanding; (2) agent response prediction; (3) relevant document identification. In our tasks, we also aim to detect the cases that are irrelevant to the associated documents, for which we modify dialogues to include \textit{irrelevant} (\texttt{Irr}) queries via data re-composition as described in Section \ref{ss:recomp}. We split the dialogues into train/dev/test sets as 70\%, 15\%, 15\% with half of the dev/test set grounded in ``unseen'' documents that are not in training set.

\begin{table*}[!t]
    \centering
    \small
    \begin{tabular}{l cccc | cccc}
    \hline
      & \multicolumn{4}{c|}{\texttt{without Irr}} & \multicolumn{4}{c}{\texttt{with Irr}} \\ 
      & \multicolumn{2}{c}{\textbf{Dev}} &  \multicolumn{2}{c|}{\textbf{Test}} & \multicolumn{2}{c}{\textbf{Dev}} &  \multicolumn{2}{c}{\textbf{Test}} \\ 
        
        \emph{\textbf{dial-ctxt}} & \textbf{EM} & \textbf{ F1} & \textbf{EM} &   \textbf{ F1} & \textbf{EM} & \textbf{ F1} & \textbf{EM} &   \textbf{ F1} \\ \hline
        
        last2 & $49.3\pm0.8$ & $64.3\pm0.3$  & $52.0\pm0.6$ & $65.5\pm0.2$  & $52.2\pm0.5$ & $63.2\pm0.1$  & $55.1\pm0.7$ & $65.3\pm0.3$   \\
all & $38.1\pm0.7$ & $52.5\pm0.5$  & $42.1\pm0.1$ & $54.3\pm0.6$  & $37.4\pm0.7$ & $44.9\pm0.7$  & $41.6\pm0.5$ & $47.7\pm0.5$   \\
last2-r & $49.4\pm1.2$ & $64.2\pm0.3$  & $52.2\pm0.4$ & $65.3\pm0.1$  & $52.1\pm1.0$ & $63.5\pm0.4$  & $55.1\pm0.4$ & $65.4\pm0.1$   \\
all-r & $50.2\pm0.5$ & $65.3\pm0.2$  & $53.1\pm0.5$ & $66.6\pm0.6$  & $52.1\pm0.9$ & $63.2\pm0.4$  & $55.4\pm0.2$ & $65.0\pm0.2$   \\
        
       \hline
    \end{tabular}
    \caption{Evaluation results for user utterance grounding. Numbers are ``mean $\pm$ stdev'' that are computed based on the results from 3 random seeds.
	}
	\label{tb:user_sp}
\end{table*}

\subsection{User Utterance Understanding}  
One of our main goals for creating this dataset is to broaden the coverage of different user queries for various task goals with respect to the associated document. Thus, our first task is interpreting a user utterance in the context of the dialogue history and the associated document. In our case, we first focus on identifying the grounding span of a user utterance.

\subsubsection{User Utterance Grounding}\label{sec-user-span-task}
In our dataset, all turns are annotated with a dialogue scene that includes the grounding span. Interpreting the user utterance could be quite challenging, as in some cases, it would depend more on the dialogue history such as U12 and U14; while in other cases, such as U1 and U16, it would depend more on current user utterance itself.
For the input of this task, it takes a user utterance along with (1) the dialogue history and 2) the document content with simplified document structure. The output is a span in the document as the reference of the given user utterance. Each grounded user turn is considered a training instance, so a dialogue with $n$ grounded user turns is considered as $n$ instances, with overlapping dialogue context.
 
\paragraph{Baseline Approach}
We formulate the problem as span selection, inspired by extractive question answering tasks such as SQuAD task~\citep{rajpurkar2016squad, rajpurkar2018know}. As a baseline, we adopt the extractive question answering model with transformers encoder by~\citet{devlin-etal-2019-bert}. 
More specifically, we follow the question answering example from \href{https://github.com/huggingface/transformers}{HuggingFace Transformers}~\citep{Wolf2019HuggingFacesTS} with pretrained bert-base-uncased model as encoder and fine-tune it during training.

The document content serves as the context input of the model. The query input is the dialogue context, for which we experiment different settings of utilizing the dialogue history: (1) last two turns (\texttt{last2}), i.e., the input user utterance for which we want to identify the dialogue scene, and the utterance before the given user utterance; (2) all previous turns (\texttt{all}), i.e., the input user utterance and all the utterances before it. We also consider different ordering type of the dialog turns: (1) in time order; and (2) in reverse order (\texttt{last2-r} and \texttt{all-r}), i.e., dialogue context is concatenated in reversed time order where the latest user utterance appears first.

Often the grounding document is longer than the maximum sequence length of transformers. In such cases, we truncate the documents in sliding windows with a stride. The dialogue context and each document trunk form one instance to be fed in batch into the encoder. The sequence of the encoded embeddings is then sent to a linear layer, which maps each embedding in the sequence into two logits, representing the probability of the corresponding position being the start and end position of the span. During training, we apply the Cross Entropy loss function to compute the loss. If the ground truth span does not fall in the document trunk, the start and end positions are both considered to be the beginning of the sequence. During decoding, the start-position and end-position logits from all document trunks are considered together to find the span most favored by the model.

\paragraph{Evaluation Metrics}
For evaluation we use Exact Match score  (\texttt{EM}) and token-level F1 score (\texttt{F1}), as in the evaluation script 2.0 of SQuAD~\href{https://worksheets.codalab.org/rest/bundles/0x6b567e1cf2e041ec80d7098f031c5c9e/contents/blob/} in Table~\ref{tb:user_sp} and Table~\ref{tb:agent_sp}.

\paragraph{Experiment Results}
The experiment results are summarized in Table~\ref{tb:user_sp}.
Generally, the model performance improves with more information added to the dialogue context. It indicates that the queries in our datasets are highly conversational contextual and our dataset could serve as a valuable source for evaluating dialogue models' capability of learning from deeper context.
We also conduct an experiment using dialogue data with \texttt{Irr}.
\texttt{Irr} turns impose noise in understanding the context, slightly reduce the model accuracy
on the original turns that are grounded to the document. However, the \texttt{Irr} turns themselves are relatively easy to identify and achieve a high score of $92.1$ with all previous turns in reverse order. As a result, the overall score of \texttt{with Irr} turns is comparable to \texttt{without Irr} in Table \ref{tb:user_sp}.

\begin{table*}[!t]
    \centering
    \small
    \begin{tabular}{l cccc | cccc}
    \hline
      & \multicolumn{4}{c|}{\texttt{without Irr}} & \multicolumn{4}{c}{\texttt{with Irr}} \\ 
      & \multicolumn{2}{c}{\textbf{Dev}} &  \multicolumn{2}{c|}{\textbf{Test}} & \multicolumn{2}{c}{\textbf{Dev}} &  \multicolumn{2}{c}{\textbf{Test}} \\ 
       
        \emph{\textbf{dial-ctxt}} & \textbf{EM} & \textbf{ F1} & \textbf{EM} &   \textbf{ F1} & \textbf{EM} & \textbf{ F1} & \textbf{EM} &   \textbf{ F1} \\ \hline
        
       last2 & $30.1\pm0.9$ & $52.1\pm0.2$  & $33.5\pm0.5$ & $52.4\pm1.1$  & $36.5\pm0.8$ & $52.8\pm0.4$  & $40.0\pm0.7$ & $53.9\pm0.6$   \\
all & $23.8\pm0.9$ & $42.4\pm0.3$  & $26.9\pm0.4$ & $42.4\pm0.2$  & $27.0\pm0.6$ & $36.4\pm0.2$  & $31.2\pm0.8$ & $39.0\pm0.3$   \\
last2-r & $29.8\pm1.2$ & $52.0\pm0.1$  & $34.2\pm0.7$ & $52.9\pm0.8$  & $35.9\pm1.3$ & $52.3\pm1.1$  & $39.9\pm1.1$ & $53.9\pm0.6$   \\
all-r & $31.1\pm1.6$ & $53.0\pm0.6$  & $34.6\pm1.1$ & $53.2\pm0.8$  & $37.2\pm0.9$ & $52.9\pm0.5$  & $41.3\pm0.8$ & $54.3\pm0.2$   \\
        
       \hline
    \end{tabular}
    \caption{Evaluation results for agent response grounding prediction. Numbers are ``mean $\pm$ stdev'' that are computed based on the results from 3 random seeds.
	}
	\label{tb:agent_sp}
\end{table*}

\subsection{Agent Response Prediction}  % Task 3.2
For this task, we aim at predicting agent responses with a focus on identifying the grounding spans in the associated document. Such kind of task can be a very important step towards building explainable conversational systems with higher practicality. In addition, we experiment with conditional text generation models to generate in-context utterance given the grounding span in the associated document.

\subsubsection{Agent Response Grounding Prediction}  
This task takes as input 1) the dialogue context; and 2) the document content with simplified document structure, and predicts a span in the document that grounds the next agent response. This task looks very similar to the user-turn grounding text prediction task in Section~\ref{sec-user-span-task} in that they both take dialogue context and document context as input and perform a span selection inside the document. However, they are essentially different: the user-turn grounding text prediction is to understand what the user has already said, whereas this task is to predict what the agent response would be based on.
 
\paragraph{Baseline Approach}
% Investigating fully the interesting yet challenging aspects of this task would involve dialogue manager and strategy modeling.
As opposed to investigating this task from the aspect of dialogue management and planning, as a first attempt, we continue with our focus on identifying the associated grounding content in the document. Thus, we treat this as a span selection task, 
% As a first attempt, we treat this as a span selection task, 
and adopt the same evaluation metrics of Exact Match scores and token level F1 scores, and the same baseline approach as in Section~\ref{sec-user-span-task}. 
Note that with the same input dialogue context and text context, the model output in Section~\ref{sec-user-span-task} is the dialogue scene corresponding to the given user utterance, 
while the model output of this task is the dialogue scene predicted for the next agent response.

\paragraph{Experiment Results}

The experiment results are summarized in Table~\ref{tb:agent_sp}.
The scores are generally much lower than the ones from our previous task in Table~\ref{tb:user_sp} due to the challenging nature of the task. We do similar trends when comparing with the experiment results for grounding user utterance task in Table~\ref{tb:user_sp}. We direct our further work on document-guided dialogue management to further improve the performance for this task.

\subsubsection{Agent Response Generation}
Next we evaluate the dataset via the task of generating agent response. This task setting considers that the span annotation is already given, then we evaluate how to generate agent utterance in context with minimized noise. Yet, this is a still quite challenging task, as in our dataset, the focused topic could be varied throughout the conversation; additionally, agent would provide either a response or follow-up ``question'' where the forms of the query turns are not restricted.

\paragraph{Baseline Approach and Experiment Results} We adopt \href{https://huggingface.co/transformers/model_doc/bart.html}{Huggingface implementation} of BART \citep{DBLP:journals/corr/abs-1910-13461}, using pretrained BART-large model as encoder, and fine-tune it during training. The input includes the user query along with dialogue history, grounding span of the next agent turn, the contexts of the grounding span and dialogue act for the next agent turn; the output is next agent utterance. For dialogue history, we consider two settings, all previous history (\texttt{all}) and last two turns (\texttt{last2}). For DA, we consider with (\texttt{+da}) and without DA in the input. We use BLEU 1-4 as evaluation of the generated utterance. For the context of the grounding span, we include the title of the document and the paragraph where the span belongs.

The BLEU scores are reported in Table \ref{tb:agent-gen}. We observe comparable scores when we include all or last two utterances. It indicates that the model might not consider the previous dialogue history much for generating agent turns. Then it might not be able to generate the reasonable answer when agents perform multi-turn verification such as D2 in Figure \ref{fig:intro}. In addition, when we add DA, the performance drops. One reason might be because writing style of the follow-up ``question'' of agent turns by human writers in our dataset, which might be quite different than questions in the pretrained models. The results confirm that even given the grounding span, generating agent response in context is a very challenging task. We leave further investigation in our future work.

\begin{table}[]
    \centering
    \small
    % \begin{tabular}{c|p{0.05\textwidth} p{0.05\textwidth} p{0.05\textwidth} p{0.05\textwidth}} \hline
    \begin{tabular}{l cccc} \hline
       \emph{\textbf{ctxt}} 
    %   &  \emph{\textbf{dial-ctxt}}
       &  \textbf{BLEU-1} & \textbf{BLEU-2} &  \textbf{BLEU-3} & \textbf{BLEU-4}  \\ \hline
last2 & 50.3 & 40.3 & 34.7 & 30.6  \\
all & 50.2 & 40.2 & 34.5 & 30.4     \\
last2+da & 48.2 & 38.1 & 32.6. & 28.6 \\
all+da  & 48.9 & 38.5 & 32.9 & 28.8 \\ \hline

% docPartial_dialPartial_woDA_wSP - 50.3/40.3/34.7/30.6
% docPartial_dialPartial_wDA_wSP - 48.2/38.1/32.6/28.6
% docPartial_dialAll_woDA_wSP - 50.2/40.2/34.5/30.4
% docPartial_dialAll_wDA_wSP - 48.9/38.5/32.9/28.8
    \end{tabular}
    \caption{Evaluation results for agent response generation with varied context input.}
    \label{tb:agent-gen}
\end{table}

\subsection{Relevant Document Identification}   % Task 1
To facilitate the understanding of the challenge of identifying the most relevant document given initial conversation turns, we next experiment with the task on identifying the grounding document given limited dialogue history information. Thus, the input is certain dialogue context and a pool of documents from all four domains.

\paragraph{Baselines and Experiment Results}
We consider two different baselines for this task: (1) BM25 \citep{RobertsonBM25} based Information Retrieval method, and (2) A multi-class sequence classifier based on multiple choice example from \href{https://github.com/huggingface/transformers}{HuggingFace Transformers}~\citep{Wolf2019HuggingFacesTS}, using pretrained bert-base-uncased model as the encoder \citep{zellers2018swag}. 

BM25 method takes the full document into account to create the index and match them against the provided dialogue contexts. BERT model takes the dialogue context $d$ and a document $y$ together as a sequence.
We use 512 tokens and feed BERT with the 256 tokens each from $d$ and $y$. 
For each dialogue context, we create a set of triples: one triple containing the correct document (labeled with $1$), and $m$ triples containing incorrect documents sampled randomly from the set of all documents (labeled with $0$). Table \ref{tb:doc_retrieval} corresponds to the setting $m=4$. During evaluation, we evaluate a given dialogue context against the set of all documents. 
%note: changed n to m, to avoid confusion with n in table caption
The task is evaluated with the commonly used recall ($R@k$) metric in retrieval tasks, which measures the fraction of times the correct document is found in the top-$k$ predictions. 

As shown in Table \ref{tb:doc_retrieval}, bert-based approach shows better performance. From the perspective of examining the quality of our dataset, we also see the numbers confirms that as more turns are included, the better the dialogue is grounded to the relevant document.

\begin{table}[]
\centering
\small
\begin{tabular}{c ccc ccc} \hline
\multicolumn{1}{l}{\multirow{2}{*}{\textbf{n}}} & \multicolumn{3}{c}{\textbf{BM-25}} & \multicolumn{3}{c}{\textbf{BERT}} \\
\multicolumn{1}{l}{} & \textbf{R@1} & \textbf{R@5} & \textbf{R@10} & \textbf{R@1} & \textbf{R@5} & \textbf{R@10} \\\hline

1 & 33.1 & 54.3 & 61.2 &  40.4 & 65.4 & 73.7 \\
2 & 57.4 & 80.0 & 84.9 &  57.8 & 81.5 & 86.2 \\
3 & 58.8 & 80.0 & 85.4 &  61.0 & 82.1 & 86.8 \\
4 & 65.4 & 85.7 & 89.9 &  63.6 & 84.9 & 89.1 \\
5 & 67.7 & 86.6 & 90.6 &  66.3 & 87.3 & 91.5 \\
 \hline
\end{tabular}
\caption{Evaluation results for document retrieval with earliest $n$ turns as input on dev set.}
    \label{tb:doc_retrieval}
\end{table}

\section{Related Work}

Our work is mainly focused on modeling dialogues that are grounded in documents. It is generally inspired by the recent substantial interests on the challenges of machine reading comprehension and conversational QA, such as CoQA~\citep{reddy2019coqa}, QuAC~\citep{choi2018quac} and DoQA~\citep{campos-etal-2020-doqa}. Those tasks aim to support conversational question answering, which involves understanding a text passage and answering a series of interconnected questions that appear in a conversation. These tasks add the complexity of co-reference resolution and contextual reasoning to the reading comprehension challenges such as SQuAD \citep{rajpurkar2016squad, rajpurkar2018know}, yet aim at identifying a solution from a given list of candidates by reasoning over spans from a document. Our task shares those challenges and additionally introduces the dialogue scenes where the agent asks questions when the user query is identified as under-specified or additional verification required for a resolute solution.

Another recent work \citet{kim2020beyond} extends MultiWOZ \citep{budzianowski2018multiwoz} by adding turns that are grounded in the FAQ knowledge for certain entity and domain. The document-based knowledge used in our work is beyond question answer pairs about certain entity of a domain but entire documents with variable contexts. In addition, ours is also largely related to conversational search tasks, such as MANtIS \citep{penha2019mantis}. Similarly, it also provides multi-turn conversations with varied user intents that are grounded in documents from Stack Exchange website. In addition to the domain difference, one major distinction is that the grounding in MANtIS is determined by the hyperlinks to a document. Our grounding is defined at at a much finer level in addition to the link to a document. 

To the best of our knowledge, the closest work to ours is ShARC \citep{saeidi2018interpretation} with dialogues that are grounded to short text snippets. It also proposes to address under-specified questions by requiring follow-up questions that are answerable with ``yes/no'' answers in similar domains. Our dataset goes beyond ShARC in several aspects nonetheless: we exploit not only paragraph-level structure but also higher-level document structure, we create conversations over much longer span of document content, where utterances are free-formed as opposed to yes/no, and we do not assume one dialogue corresponds to one single goal.

\section{Conclusion}
We have introduced \textbf{\texttt{doc2dial}}, a new dialogue dataset for goal-oriented tasks that are grounded in documents from multiple domains. Compared to previous work, our dialogues cover a greater variety of dialogue scenes that correspond to both semi-structured and unstructured contents with a much wider span in the associated documents. For evaluation, we investigated three types of dialogue tasks and proposed baseline approaches. We hope this work will inspire and assist both dialogue and document modeling for tackling more goal-oriented dialogue tasks in practice.

\bibliographystyle{acl_natbib}
\bibliography{emnlp2020}

\end{document}